\documentclass[runningheads]{llncs}
\usepackage{graphicx}
\usepackage{amsmath,amssymb} 
\usepackage{color}
\usepackage{adjustbox,multirow}

\begin{document}

\title{Detector-in-Detector: Multi-Level Analysis for Human-Parts} 
\titlerunning{DIDNet} 


\author{Xiaojie Li\inst{1}\orcidID{0000-0001-6449-2727} \and
Lu Yang\inst{2}\orcidID{0000-0003-3857-3982} \and
Qing Song\inst{2}\orcidID{0000−0003−4616−2200} \and
Fuqiang Zhou\inst{1}\orcidID{0000-0001-9341-9342}}
%

\authorrunning{L. Xiaojie, Y. Lu et al.} 


\institute{Beihang University, Beijing 100191, China \and
Beijing University of Posts and Telecommunications Beijing 100876, China \\
\email{\{xiaojieli,zfq\}@buaa.edu.cn} \\
\email{\{soeaver,songqing512\}@bupt.edu.cn}}

\maketitle

\begin{abstract}
Vision-based person, hand or face detection approaches have achieved incredible success in recent years with the development of deep convolutional neural network (CNN). In this paper, we take the inherent correlation between the body and body parts into account and propose a new framework to boost up the detection performance of the multi-level objects. In particular, we adopt region-based object detection structure with two carefully designed detectors to separately pay attention to the human body and body parts in a coarse-to-fine manner, which we call Detector-in-Detector network (DID-Net). The first detector is designed to detect human body, hand and face. The second detector, based on the body detection results of the first detector, mainly focus on detection of small hand and face inside each body. The framework is trained in an end-to-end way by optimizing a multi-task loss. Due to the lack of human body, face and hand detection dataset, we have collected and labeled a new large dataset named \textit{Human-Parts} with 14,962 images and 106,879 annotations. Experiments show that our method can achieve excellent performance on \textit{Human-Parts}.

\keywords{Convolutional neural network \and Detector in Detector \and Human parts.}
\end{abstract}

\section{Introduction}
\label{sec:intro}
Robust detection of human body, face and hand in the wild are canonical sub-problems of general object detection. They are prerequisite for various person-based tasks such as pedestrian detection~\cite{Ribeiro2016A}, person re-identification~\cite{Chen2017Cascaded,Papandreou2017Towards}, facial landmarking~\cite{Xiao2017Recurrent} and driver behavior monitoring~\cite{Dhawan2013Implementation}. The problems of human parts detection in the wild have been intensely studied for decades and significant progress have been made in recent detection algorithms due to the advancement of deep Convolutional Neural Networks (CNN) such as \cite{Ribeiro2016A,Ghorban2018Aggregated} in person detection, \cite{Samangouei2018Face,Zhu2017CMS} in face detection and \cite{Deng2016Joint,Le2017Robust,Mittal2011Hand} in hand detection.

\begin{figure}
	\centering
	\includegraphics[width=1\textwidth]{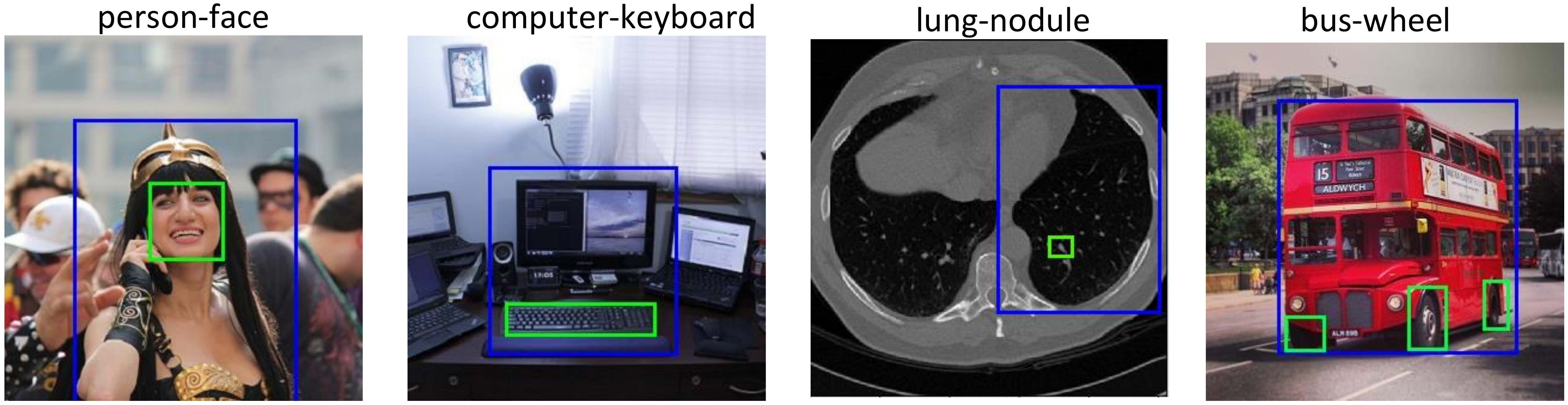}
	\caption{Examples of multi-level objects. Boxes in green are sub-objects of boxes in blue.} 
	\label{fig:intro}
\end{figure}

Human parts are multi-level objects~\cite{Zhao2010Semantic}, where face and hand are sub-objects of body. There are many other multi-level objects in our daily life such as laptop and keyboard, lung and lung-nodule or bus and wheel, which are shown in Fig.~\ref{fig:intro}. However, most detection frameworks ignore the inherent correlation between multi-level objects and coarsely treat these \textbf{sub-objects} and \textbf{objects} as normal objects when they solve this multi-level objects detection problem. In this paper, we perform the person, face and hand detection tasks together to explore the more efficient detection methods for the multi-level objects.

When doing this multi-level objects task using general detection algorithm, detection performance for large objects, such as the human body, is relatively straightforward. The crucial challenges in real-world applications mainly come from training detectors for small objects such as face and hand due to large pose variations and serious occlusions, which make them still far from achieving the same detection capabilities as a human. Due to the large scale variance between the body (objects) and small body parts (sub-objects), the whole image is mainly occupied by the big objects like human body. Small hand and face usually occupy a relatively smaller area in practice. Thus, there are more background information than the small objects during training, which results in a serious disturbance when doing small objects detection.
To cope with the problems, inspired by the top-down pose estimation approaches that first locate and crop all persons from image and then solve the single person pose estimation problem in the cropped person patches~\cite{Papandreou2017Towards,Chen2017Cascaded}, we proposed a region-based convolutional neural network, named Detector-in-Detector Network (\textbf{DID-Net}), which allows the network to see the body in the pictures first and then look inside these bodies to find the tiny faces and hands. 
Our network contains two region-based detectors: \textit{BodyDetector} and \textit{PartsDetector}. The \textit{BodyDetector} adopt the traditional Faster R-CNN implementation to predict a set of human body bounding boxes. Then the feature maps of the detected bodies are wrapped and sent to the second region-based \textit{PartsDetector} to predict the bounding boxes of hand and face. In \textit{PartsDetector}, many background regions that are useless for training are cut down. Thus there is less disturbance inside the cropped body features, which is beneficial to the detection of small parts. The whole network is trained in an end-to-end way.

In order to demonstrate the proposed method in more practical scenes, we construct a new \textit{Human-Parts} dataset, which contains 14,962 well-labeled images for human body, face and hand in realistic unconstrained conditions. To the best of our knowledge, it is the first detection dataset that combines three important parts of human. The dataset is available at \url{https://github.com/xiaojie1017/
Human-Parts}.

\section{Related Work}
\label{sec:related}
\subsection{Object Detection methods}
CNN-based object detection algorithms fall into two categories. The first category is built upon a two-stage pipeline~\cite{Girshick2015Fast,Ren2015Faster,Lin2016Feature}, where the first stage proposes candidate object bounding boxes and the second stage extracts features using RoIPool from each candidate box and performs classification and bounding-box regression. The second category divides the image into a grid, then simultaneously makes prediction for each square or rectangle in the grid, and finally figures out the bounding boxes of targeting objects based on the predictions of the squares or rectangle, such as \cite{Liu2015SSD,Li2017FSSD,Redmon2015You}. That design will achieve relatively faster computational speed. However, the one-stage methods can not achieve comparable accuracy with two-stage approaches. In this regard, most of the existing human parts detection methods employ the two-stage pipeline such as~\cite{Zhu2017CMS,Jiang2016Face,He2017A}.
\subsection{Human-Parts detection methods}
Faster R-CNN~\cite{Ren2015Faster}  trains the object proposal and classifier at the same time on the same base network. Region Proposal Networks (RPN) in the Faster R-CNN~\cite{Ren2015Faster} are successful in eliminating the need for a precomputed object proposals. \cite{Jiang2016Face,He2017A,Zhu2017CMS} as well as our work are all the frameworks extended from Faster R-CNN framework. Other object-proposal-free detectors for human parts such as \cite{Hu2016Finding,Najibi2017SSH}, perform detection in a fully convolutional manner as RPN does for proposing bounding boxes. Cascaded stages design algorithms are widely used in human parts detection tasks recently such as~\cite{Li2015A,Zhang2016Joint}. The advantage of these cascaded stages lies in that they can handle unbalanced distribution of negative and positive samples. In the low-resolution stages, weak classifiers can reject most false positives. In the high-resolution stages, stronger classifiers can save computation with fewer proposals. \cite{Li2015A} employs a cascade of classifiers to efficiently reject backgrounds. In~\cite{Zhang2016Joint} three stages of carefully designed CNNs are used to predict face and landmark in a coarse-to-fine manner. However, most of detection methods perform multi-level objects detection together without taking the inherent correlation between them into account.
\subsection{Top-down pose estimation methods}
Top-down human pose estimation algorithms perform pose estimation task using two separated networks: one person detection network and one single person pose estimation network. Persons are first detected in an image with the detector. Then the detected person will be cropped from the image and the single person pose estimation network is used to predict the keypoints of each person. A number of top-down algorithms achieve excellent performance on COCO~\cite{Lin2014Microsoft} keypoint benchmark such as~\cite{Chen2017Cascaded,Papandreou2017Towards}. By sharing features efficiently, Mask-RCNN~\cite{He2017Mask} can do this task in an end-to-end approach. It first generates proposals of person. Then it predict $K$ masks for each proposal, one for each of \textit{K} keypoint types, which is a Pose-estimator-in-Detector structure.

As a consequence, instead of conducting traditional cascaded stages detection methods that refine predictions step by step, we deal with the multi-level objects detection task using a Detector-in-Detector network. That approach perform parts detection inside each person proposal inspired by the top-down pose estimation methods. Details will be discussed in Section~\ref{sec:methods}.
\section{Detector-in-Detector Network}
\label{sec:methods}
\begin{figure}
	\centering
	\includegraphics[width=1\linewidth]{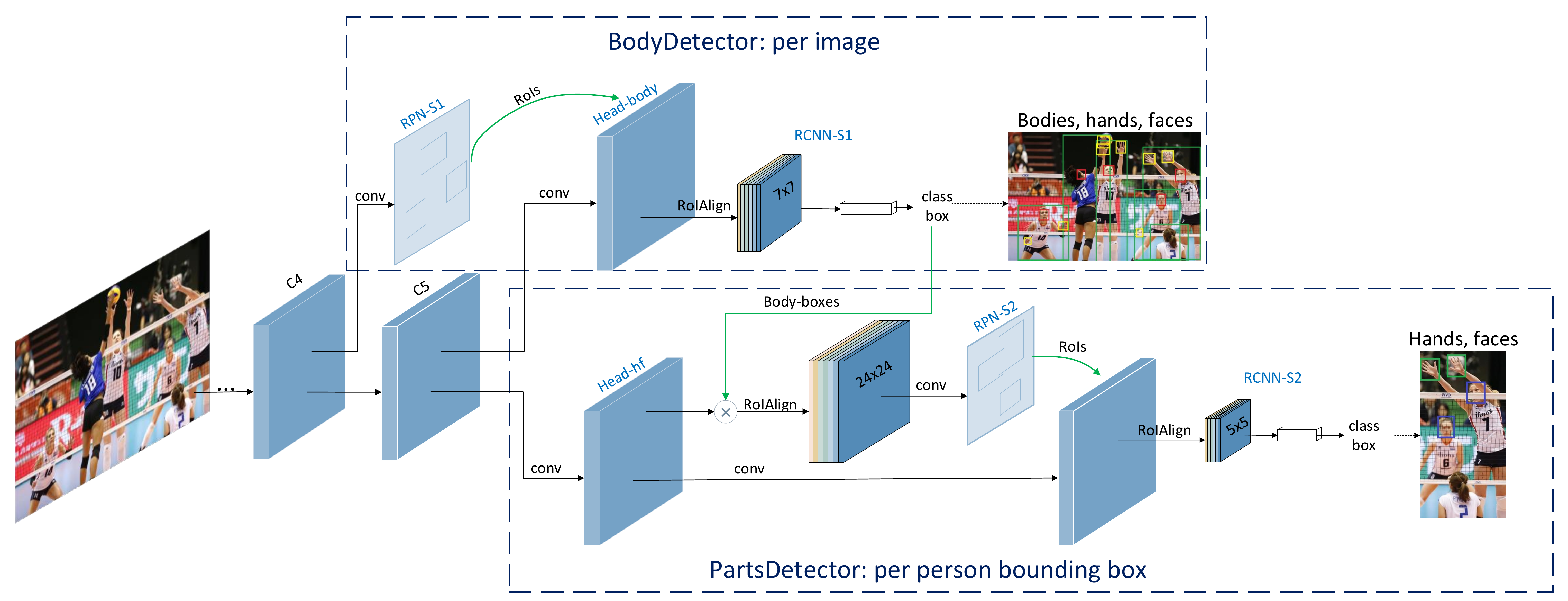}
	\caption{The pipeline of the proposed DID-Net framework. (1)\textit{BodyDetector}, which contains RPN-S1 and RCNN-S1, is used to detect body, hand and face. (2)\textit{BodyDetector}, which contains RPN-S2 and RCNN-S2, is used to predict hand and face inside the body bounding boxes detected by \textit{BodyDetector}.}
	\label{fig:framework}
\end{figure}

The architecture of the proposed Detector-in-Detector network (DID-Net) is illustrated in Fig.~\ref{fig:framework}. We extended Faster RCNN structure by appending another light-weight parts detector after it, which is a Detertor-in-Detector structure. The framework contains two detectors: (1) \textit{BodyDetector} is designed for human body, hand and face detection. (2) \textit{PartsDetector} is appended for hand and face detection inside the selected body bounding boxes, which are detected by \textit{BodyDetector}. 

\subsection{BodyDetector}
\textit{BodyDetector} uses the same structure with Faster RCNN. Our backbone network is Resnet-50~\cite{He2016Deep}, which is pre-trainied on ImageNet~\cite{Deng2009ImageNet}. First, the backbone network generates a series of convolutional feature maps at several scales. We denote these feature maps as (C2, C3, C4, C5) which is extracted from last layer of each scale stage. To increase the resolution of the C5 feature maps, we reduce the effective stride from 32 pixels to 16 pixels as done in~\cite{Dai2016R}. Consequently, their corresponding strides of each layer are (4, 8, 16, 16) pixels with respect to the input image. C4 features are used to predict a set of proposals using the region proposal network, which we named it RPN-S1. Then features of each proposal are cropped from Head-body and pooled into a fixed-size feature vector using RoI Align (We use $7\times7$ size as \cite{Ren2015Faster} does). The features are then fed into RCNN-S1 to do per-proposal classification operation and box refinement of body, hand and face. For model compactness, Head-body is generated after a $1\times 1$ convolutional layer with 256 kernels over the top features of C5 to reduce the dimension. The RoI Align layer is added on the top of the feature maps of Head-body. Two new fully connected (fc) layers of dimension $1024$ are applied on the RoI Align features, followed by the bounding box regression and classification branches.

\subsection{PartsDetector}
\textit{PartsDetector} shares the same convolutional features of the first stage by simply adding the \textbf{Head-hf} to C5 in parallel with \textbf{Head-body}, which contains 256 channels features as the same as \textbf{Head-body}. The predicted body bounding boxes of the first stage with high classification scores will be chosen. Non-maximum suppression (NMS)~\cite{Girshick2014Deformable} is performed on the chosen proposals to eliminate highly overlapped detection bounding boxes. After box regression, the features of person proposals will be wrapped from Head-hf with RoI Align and pooled into size $24\times 24$. RPN-S2 is applied on the wrapped body features to generate RoIs of hand and face. Then the features of RoIs will be extracted from another set of features, which are 256 channels features outputted by adding two $3\times 3$ convolutional layers over Head-hf. Two fully connected layers are appended after the extracted RoI features for further classification and box regression using RCNN-S2. In this stage, we use size $24\times 24$ for the RoI Align size of body and use size $5\times 5$ for the cropped features of hand or face due to the scale variance between them.

During training, we select the top 16 body detections within each image considering the dataset we used, which usually contains a maximum number of 11 persons in one image. While the positions of body bounding boxes are generated during training and are uncertain before. To handle this, we generate the ground truth annotations inside the predicted human body in an online way. When the original face and hand ground truth lie inside or partly inside the detected body bounding boxes, They will be regarded as the ground truth annotations of that body. After transferring the coordinate of these hand and face boxes from the original image to the body bounding boxes, a new batch of training data for hand and face inside each body can be sent to the \textit{PartsDetector} for parts detection. In this detector, we treat every detected human body as one input image in the first detector when writing the implementation code. NMS is applied to eliminate highly overlapped detection bounding boxes during inference and training.

\subsection{Loss function}
Our network is trained to minimize a multi-task loss, which is composed of classification and bounding-box regression losses of RPN and RCNN modules from the two detectors. We use softmax cross-entropy loss for classification and smooth \textit{L1} loss~\cite{Ren2015Faster} for bounding-box regression. For the RPN stage in two detectors, the classification loss is softmax cross-entropy loss over background/foreground classes. The loss can be formulated as Equation~\ref{equ:total_loss}. Here, $\mathcal{L}_{rpn}^{s1}$ and $\mathcal{L}_{cls}^{s1}$ denote the losses of RPN-S1 and RCNN-S1 in \textit{BodyDetector}, and $\mathcal{L}_{rpn}^{s2}$ and $\mathcal{L}_{cls}^{s2}$ denote losses of RPN-S1 and RCNN-S2 in \textit{PartsDetector}.

\begin{equation}
\label{equ:total_loss}
\mathcal{L} = \mathcal{L}_{rpn}^{s1} + \mathcal{L}_{cls}^{s1} + \mathcal{L}_{rpn}^{s2} + \mathcal{L}_{cls}^{s2}
\end{equation}

\section{Experiments}
\label{sec:experiments}

\subsection{Human-Parts Datasets}
\label{subsec:dataset}
\begin{figure}
	\centering
	\includegraphics[width=1\linewidth]{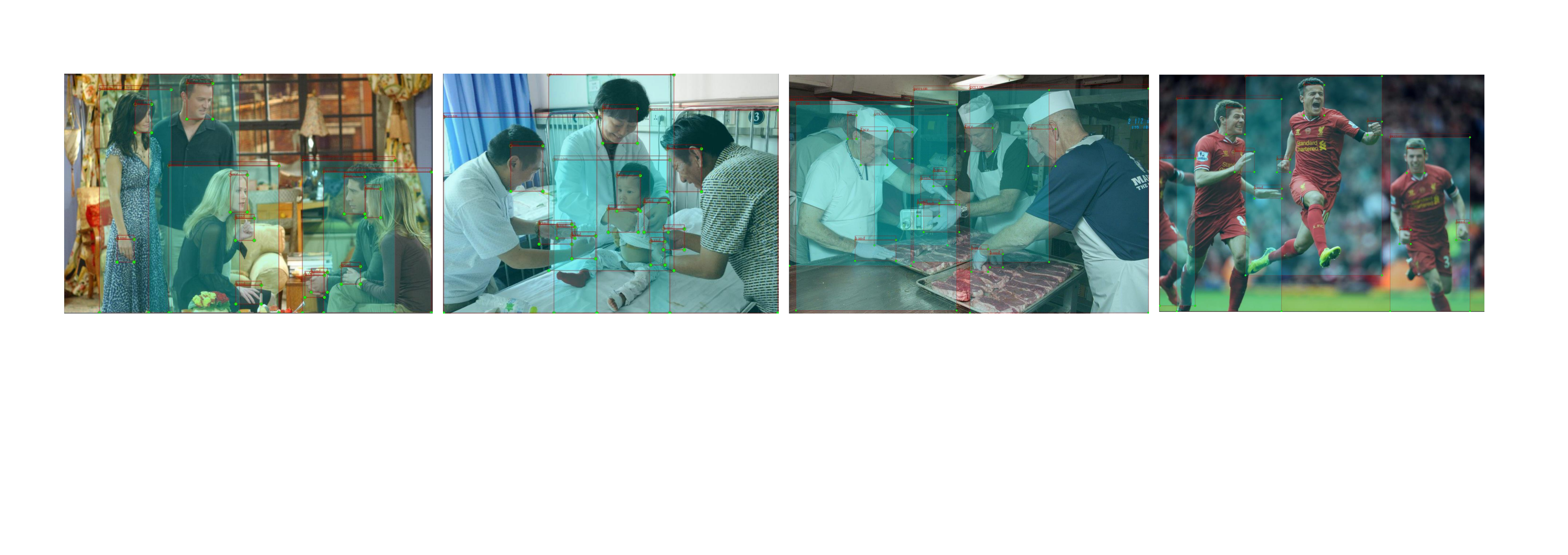}
	\caption{Samples of annotated images in \textit{Human-Parts} dataset.}
	\label{fig:dataset}
\end{figure}
In this paper, we collected and labeled a detection dataset named \textit{Human-Parts} which contains annotations of three categories, including person, hand and face. The proposed dataset contains high-resolution images which are randomly selected from AI-challenger~\cite{Wu2017AI} dataset. Person category has already been labeled in this dataset. However, the small human whose body parts are hard to distinguish or the vague ones whose body contours are hard to recognize are missed-labeled in this dataset. We added the missed person body annotations and labeled hand and face additionally in each image. The number of persons in each image range from 1 to 11. In total, our dataset consists of 14,962 images (we use 12,000 for train, 2,962 for testing) with 10,6879 annotations (35,306 persons, 27,821 faces and 43,752 hands). Our dataset followed the same standard annotation principles as Wider Face~\cite{Yang2016WIDER} and VGG Hand dataset~\cite{Mittal2011Hand}. We have labeled every visible person, hand or face with xmin, ymin, xmax and ymax coordinates and ensured that annotations cover the entire objects including the blocked parts but without extra background. The annotation format follows PASCAL VOC~\cite{Everingham2010The}. The representative images and annotations are shown in Fig~\ref{fig:dataset}. More details about this dataset and comparisons with other human parts datasets can be seen in Table~\ref{tab:dataset}.
\begin{table*}
	\caption{Comparison of different human parts detection datasets}
	\label{tab:dataset}
	\begin{center}
		\begin{adjustbox}{max width=0.8\textwidth}
			\begin{tabular}{c|c|ccc|c}
				\hline
				DataSet                     & Images &   Person   &    Hand    &    Face    & Total Instance \\ \hline\hline
				\emph{Caltech}~\cite{Dollar2009Pedestrian}   & 42,782 & \checkmark &     -      &     -      &     13,674     \\
				\emph{CityPersons}~\cite{Zhang2017CityPersons} & 2,975  & \checkmark &     -      &     -      &     19,238     \\ \hline
				\emph{VGG Hand}~\cite{Mittal2011Hand}      & 4,800  &     -      & \checkmark &     -      &     15,053     \\
				\emph{EgoHands}~\cite{Huang2017Egocentric}   & 11,194 &     -      & \checkmark &     -      &     13,050     \\ \hline
				\emph{FDDB}~\cite{Jain2010FDDB}         & 2,854  &     -      &     -      & \checkmark &     5,171      \\
				\emph{Wider Face}~\cite{Yang2016WIDER}     & 32,203 &     -      &     -      & \checkmark &    393,703     \\ \hline
				\emph{Human Parts}               & 14,962 & \checkmark & \checkmark & \checkmark &    106,879     \\ \hline
			\end{tabular}
		\end{adjustbox}
	\end{center}
\end{table*}

\subsection{Implementation Details}	
\label{subsec:details}
We perform all experiments on \textit{Human-Parts} dataset. Training and evaluation are performed on the 12,000 images in the \textit{train} set and the 2,962 images on the \textit{test} set. For evaluation, we use the standard average precision(AP) and mean average precision(mAP). We report AP and mAP scores using the intersection over union (IoU)~\cite{Everingham2010The} threshold at 0.5. All networks are fine-tuned from a pre-trained ImageNet classification network ResNet-50. Our system is implemented in Pytorch and source code will be made publicly available. 

For anchor generation, we use scales ($64^2$, $128^2$, $256^2$, $512^2$) in the \textit{BodyDetector} stage and ($32^2$, $64^2$, $128^2$, $256^2$) in the \textit{PartsDetector} stage considering the scale variance of body and body parts. All anchors have the aspect ratio of (0.5, 1, 2) due to the large variations of hand and face in the wild. During training, the mini-batch size of 512 is employed for the RCNN-S1 stage of \textit{BodyDetector}, and 32 is employed for each person RoI in the RCNN-S2 stage of \textit{PartsDetector}. Multi-scale training strategy is adopted to be robust to different scale objects. In our method, the shorter side is resized to (416, 480, 576, 688, 864, 1024) pixels. We only use horizontal image flipping augmentation. In the testing phase, the shorter side of each image is resized to 800 pixels and tested independently. For other settings, we follows the work \cite{Ren2015Faster}.

During training, we adopt an end-to-end learning procedure using stochastic gradient descent (SGD). The whole network is trained for 30,000 iterations with initial learning rate of 0.01. We decay the learning rate by 0.1 at 20,000 iterations. A momentum of 0.9 and a weight decay of 0.0005 are used. During inference, \textit{BodyDetector} module outputs 300 best scoring anchors as detections. While in the \textit{PartsDetector} module, we only select 30 best scoring anchors on account of limited parts inside each body region. The results of hand and face in \textit{BodyDetector} and \textit{PartsDetector} are fused together and NMS with a threshold of 0.45 is performed on the outputs due to the existence of large overlap objects in the wild such as Fig.~\ref{fig:dataset}.

\subsection{Main results}
\label{subsec:main_results}
In Table~\ref{tab:baseline}, we perform studies of traditional Faster RCNN training with different data and our DID-Net. \emph{Separate Network} adopts only one category detection task in one ResNet-50 Faster R-CNN network. \emph{Union Network} performs three categories detection together in the same Network as \emph{Single Network}. From Table~\ref{tab:baseline} we can see that: 
\begin{table*}
	\caption{Basic results of Faster RCNN and our DID-Net. \textbf{P} denotes person, \textbf{H} denotes hand and \textbf{F} denote face. The bold values are the best performance in each column}
	\label{tab:baseline}
	\begin{center}
		\begin{tabular}{c|c|c|ccc|c}
			\hline
			& Train Data &             Detector              &   Person AP   &    Face AP    &    Hand AP    &          mAP          \\ \hline\hline
			\multirow{3}{*}{\emph{Separate Network}} &     P      & \multirow{3}{*}{ResNet-50-Faster} &     88.1      &       -       &       -       & \multirow{3}{*}{90.2} \\
			&     F      &                                   &       -       &     95.9      &       -       &                       \\
			&     H      &                                   &       -       &       -       &     86.7      &                       \\ \hline
			\emph{Union Network}           & P + H + F  &         ResNet-50 -Faster         &     89.3      &     93.0      &     82.3      &         88.2          \\ \hline
			\emph{Our Network}            & P + H + F  &           ResNet-50-DID           & \textbf{89.6} & \textbf{96.1} & \textbf{87.5} &     \textbf{91.1}     \\ \hline
		\end{tabular}
	\end{center}
\end{table*}

\begin{itemize}
	\item \textbf{Person AP} The AP performance of person is increased from $88.1$ to $89.6$ after conducting multi-objects detection task together in a single network, from which we can infer that the saliency of hand and face will contribute to the detection performance of the person. Works of ~\cite{Qin2016Joint,Yang2017Faceness} also show that the facial contributes based supervision can effectively enhance the capability of a face detection network. 
	\item \textbf{Small Parts AP} The AP values of small parts in \emph{Union Nework} decrease compared with the \emph{Separate Network}. The reasons mainly come from the scale variance between human body and body parts. In practice, the whole image is mainly occupied by the big objects like human body. Small hand and face usually occupy a relatively smaller area. Given a fixed anchor scales and anchor ratios, there will be less small anchors generated by RPN, which lead to the poor performance of small parts detection.
\end{itemize}

Compared with \emph{Single Network} and \emph{Union Network}, DID-Net conducts another RPN inside each body, where less disturbance from other objects or background exist. The network can learn more spacial region relationship between body parts. In addition, when extracting features of the person proposals, we pooled these wrapped features to size $24\times24$, which will be larger than the original person feature maps mostly. That operation will zoom out the features of the small parts and result in a high resolution of feature maps for further detection. Results show that our DID-Net outperforms the \emph{Separate Network} and \emph{Union Network}, which demonstrate the efficiency of our framework.

\subsection{Ablation study}
\label{subsec:ablation_study}
In Table~\ref{tab:partsdetector}, we evaluate how different components affect the detection performance of \textit{PartsDetector}. For fair comparison, we fixed the parameters of the backbone network and \textit{BodyDetector}. Thus the person AP in Table~\ref{tab:partsdetector} is constant.
\begin{table}[htb!]
	\caption{Experiment results about how structure design and the number of training bodies affect the detection performance of \textit{PartsDetector}}
	\label{tab:partsdetector}
	\begin{center}
		\begin{tabular}{c|c|ccc|c}
			\hline
			\textit{PartsDetector}  & Person RoI & Person AP &    Face AP    &    Hand AP    &      mAP      \\ \hline\hline
			\emph{RPN-S2 + RCNN-S2} &     16     &   89.6    & \textbf{96.1} & \textbf{87.5} & \textbf{91.1} \\ \hline
			\emph{RPN-S2}      &     16     &   89.6    &     93.5      &     83.3      &     88.8      \\ \hline
			\emph{RPN-S2 + RCNN-S2} &     8      &   89.6    &     95.4      &     86.0      &     90.3      \\ \hline
		\end{tabular}
	\end{center}
\end{table}
\begin{itemize}
	\item \textbf{Architecture.} There are two designs of \textit{PartsDetector}: (1) \textit{RPN-S2 + RCNN-S2} design: It follows the structure we described in Section \ref{sec:methods}. Two RoI Align operations are adopted as shown in Fig.~\ref{fig:framework}. (2) \textit{RPN-S2} is changed to output the classification and box regression of each RoI directly as SSD did. There is only one RoI Align operation. Table~\ref{tab:partsdetector} shows that two-stage design of \textit{PartsDetector} can achieve a higher accuracy on small parts detection.
	\item \textbf{Person RoI.} Person RoI represents that how many person bodies after NMS opetation are selected from the \textit{BodyDetector} during training. We use a different number of $8$ and $16$ for this number and conduct experiments. Larger number is not used considering the computation efficiency. Results show that, if the selected bodies are less than the bodies the image really contains during training, the features of Head-hf will not contain enough response to unselected bodies. And that incomplete training will lead to a performance drop of the \textit{PartsDetector}.
\end{itemize}

\subsection{Comparisons with the state-of-arts}
\label{subsec:comparions}

\begin{table*}[htb!]
	\caption{Table of Average Precision on validation set of Human parts. Our DID-Net outperforms SSD~\cite{Liu2015SSD}, Faster R-CNN~\cite{Ren2015Faster}, RFCN~\cite{Dai2016R} with OHEM~\cite{Shrivastava2016Training} and FPN~\cite{Lin2016Feature} on person and face detection. Comparable results are achieved on face detection ability compared with FPN}
	\label{tab:state-of-art}
	\begin{center}
		\begin{adjustbox}{max width=0.95\textwidth}
			\begin{tabular}{c|c|c|c|c|c|c|c}
				\hline
				& backbone  & Multi scale & RoI Align  & $AP_{person}$ &  $AP_{face}$  &  $AP_{hand}$  &      mAP      \\ \hline\hline
				SSD        &   VGG16   &      -      &     -      &     84.3      &     90.4      &     77.4      &     84.0      \\
				Faster R-CNN   & ResNet-50 & \checkmark  & \checkmark &     89.2      &     93.0      &     82.3      &     88.1      \\
				RFCN + ohem    & ResNet-50 & \checkmark  & \checkmark &     88.9      &     93.2      &     84.5      &     88.9      \\
				FPN        & ResNet-50 & \checkmark  & \checkmark &     87.9      & \textbf{96.5} &     85.4      &     89.9      \\ \hline
				\textbf{DID-Net} & ResNet-50 & \checkmark  & \checkmark & \textbf{89.6} &     96.1      & \textbf{87.5} & \textbf{91.1} \\ \hline
			\end{tabular}
		\end{adjustbox}
	\end{center}
\end{table*}

We compare the proposed DID-Net to the state-of-the-art methods performed on \textit{Human-Parts} dataset in Table~\ref{tab:state-of-art}, which includes SSD (training image size are $512\times512$), Faster R-CNN, RFCN~\cite{Dai2016R} with online hard example mining(OHEM)~\cite{Shrivastava2016Training} and FPN~\cite{Lin2016Feature}. SSD is conducted using original settings of paper. For those region-based detectors, multi-scale training (described in \ref{subsec:details}), anchor scales of ($64^2$, $128^2$, $256^2$, $512^2$), anchor ratios of (0.5, 1, 2) and RoI Align are adopted in all models during training. All the models are trained for 30,000 iterations with a initial learning rate of 0.01, a weight decay of 0.0005 and a momentum of 0.9.

Results show that our DID-Net achieves the highest accuracy on person and hand category. Region-based two-stage detection models (Faster R-CNN, RFCN, FPN) outperform single-stage detectors (SSD), which demonstrates the ability of two-stage detectors. FPN achieves better performance on hand and face detection performance than Faster R-CNN owing to the efficient fusion of features with different scales. While the performance of person dropped slightly than Faster R-CNN due to the scale variance between human body and body parts. After adding the \textit{PartsDetector} to the basic \textit{BodyDetector}, the detection performance of  face and hand in our architecture have a large promotion.

\subsection{Qualitative Results}
\begin{figure*}[htb!]
	\centering
	\includegraphics[width=0.95\linewidth]{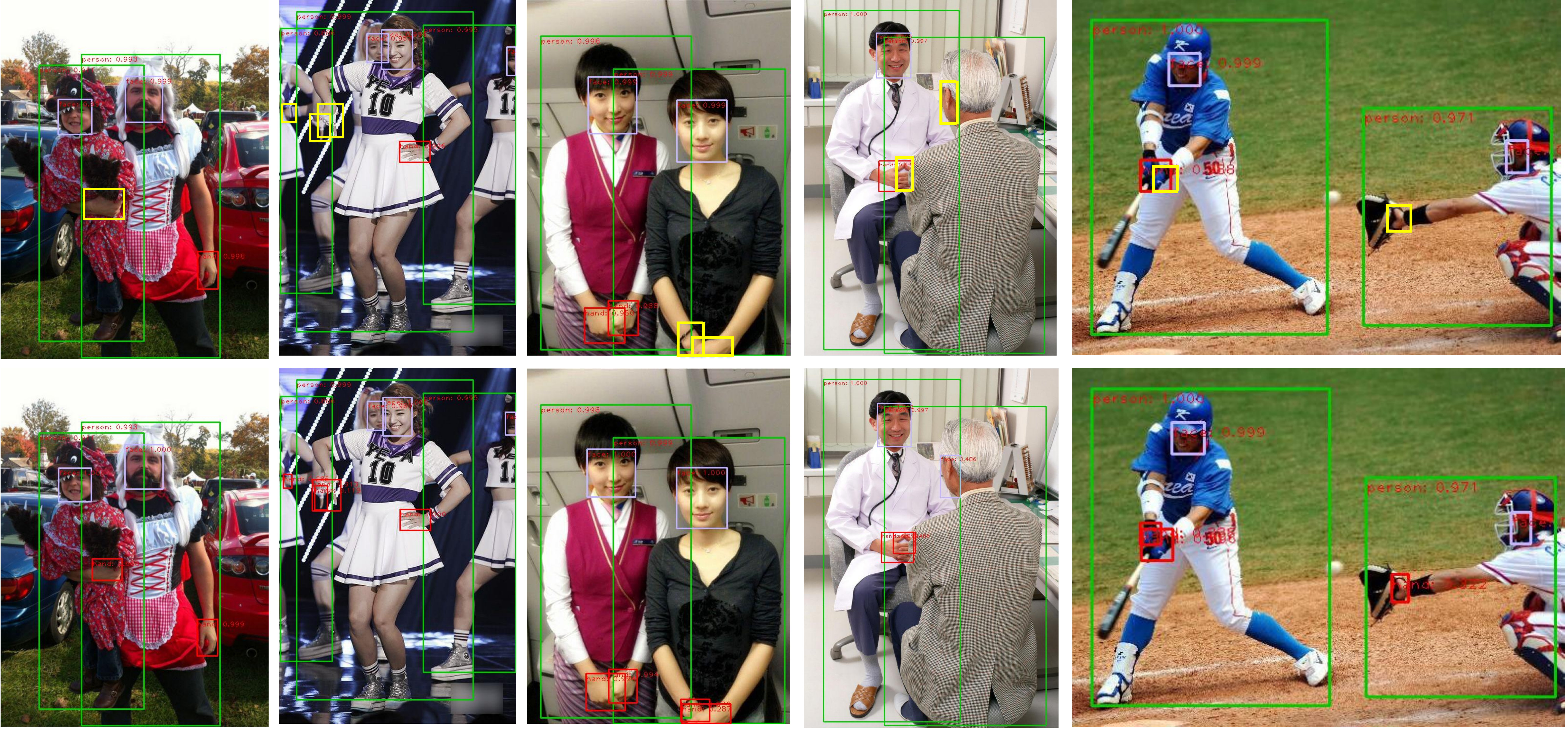}
	\caption{Some example results of Faster RCNN (Top) and the proposed DID-Net (Bottom). Green boxes are the detected persons. Red boxes are the detected hands. Purple boxes are the detected faces. Yellow boxes in the first row are the missed objects of Faster-RCNN.}
	\label{fig:results}
\end{figure*}
We show some qualitative human parts detection results on sample images in Fig.~\ref{fig:results}. From the Faster RCNN results in the first row, we observe that there still exists some small hands or faces (in yellow boxes), which can not be detected. While in the results of DID-Net, those hard parts can be found. Results can be seen in the second row of Fig.~\ref{fig:results}. 

\section{Conclusion}
\label{sec:conclusion}
In this paper, we propose a Detector-in-Detector network (DID-Net) for the multi-level objects by simply appending a light-weight \textit{PartsDetector} after Faster RCNN structure to perform parts detection inside each body. We implement our methods on \textit{Human-Parts} dataset and experiments show that our methods can achieve an excellent performance, especially for small hands and faces. The novel framework we constructed can also be performed on other multi-level objects. In addition, we have also build a dataset named \textit{Human-Parts}, which aims to the human body, hand, and face in many realistic environments. Others can also use our dataset for related tasks and applications.

\end{document}